\documentclass[11pt,a4paper,table]{article}
   \makeatletter
\def\@fnsymbol#1{\ensuremath{\ifcase#1\or \dagger\or \ddagger\or 
    \mathsection\or \mathparagraph\or \|\or **\or \dagger\dagger
    \or \ddagger\ddagger \else\@ctrerr\fi}}
    \makeatother
\usepackage[hyperref]{eacl2021}
\usepackage{times}
\usepackage{latexsym}

\usepackage[font=small,skip=10pt]{caption}
\usepackage{microtype}
\usepackage{todonotes}
\usepackage{graphicx}
\usepackage{color, colortbl}
\definecolor{Gray}{gray}{0.3}
\usepackage{xcolor}
\usepackage{multirow}
\usepackage{float}
\restylefloat{table}

\aclfinalcopy % Uncomment this line for the final submission

\title{Beyond the English Web: Zero-Shot Cross-Lingual and\\ Lightweight Monolingual Classification of Registers}

\author{Liina Repo$^{\ast}$\thanks{~~The marked authors contributed equally to this paper.}~~~Valtteri Skantsi$^{\ast\circ}$\footnotemark[1]~~~Samuel Rönnqvist$^{\ast}$~~~Saara Hellström$^{\ast}$\\ 
{\bf \large Miika Oinonen$^{\ast}$~~~Anna Salmela$^{\ast}$~~~Douglas Biber$^{\ddagger}$~~~Jesse Egbert$^{\ddagger}$}\\
{\bf \large Sampo Pyysalo$^{\ast}$~~~Veronika Laippala$^{\ast}$}\\[2mm]
$^{\ast}$University of Turku~~~$^{\circ}$University of Oulu~~~$^{\ddagger}$Northern Arizona University\\[1mm]
$^{\ast}$\{tlkrep,valtteri.skantsi,saanro,sherik,mhtoin,annsaln,sampo.pyysalo,mavela\}@utu.fi\\
$^{\circ}$\{valtteri.skantsi\}@oulu.fi~~~$^{\ddagger}$\{douglas.biber,jesse.egbert\}@nau.edu}

\date{}

\begin{document}
\maketitle
\begin{abstract}

We explore cross-lingual transfer of register classification for web documents. Registers, that is, text varieties such as blogs or news are one of the primary predictors of linguistic variation and thus affect the automatic processing of language. We introduce two new register-annotated corpora, FreCORE and SweCORE, for French and Swedish. We demonstrate that deep pre-trained language models perform strongly in these languages and outperform previous state-of-the-art in English and Finnish. Specifically, we show 1) that zero-shot cross-lingual transfer from the large English CORE corpus can match or surpass previously published monolingual models, and 2) that lightweight monolingual classification requiring very little training data can reach or surpass our zero-shot performance. We further analyse classification results finding that certain registers continue to pose challenges in particular for cross-lingual transfer. 
\end{abstract}

\section{Introduction}
\label{sec:intro}

Text genre or \textit{register} \cite{biber:1988}, such as discussion forum, news article or poem, is one of the most important predictors of linguistic variation \cite{biber2012-predictor}. Thus, register affects crucially also the automatic processing of language \cite{mahajan-2015-feature,webber2009genre,van2018evaluation}. Yet, despite its importance, register information is not available in web-crawled datasets that are widely used e.g.\ for pre-training language models in modern NLP. This is a challenge, as better structured language resources would also enable more detailed understanding and more sophisticated use of this data.

While web register identification would allow better realization of the potential offered by web-crawled datasets, most previous web register identification studies have been limited by skewed datasets, low performance, and near-exclusive focus on English. For example, \newcite{asheghi2014} and \newcite{Pritsos2018} reported comparatively strong results, but their evaluations were based on datasets representing only a subset of the registers found online. With the CORE corpus, \newcite{egbert:core} were the first to present a dataset featuring the full extent of registers found on the open, searchable English web. While \newcite{biber2016-detecting} demonstrated the possibility of automatic register classification using Stepwise Discriminant Analysis, improvements in modeling and more efficient methods remained necessary in order to reach practical levels of performance. 

A challenge in modeling web registers is that web documents drawn from the unrestricted web do not always fit discrete classes but could rather be described in a continuous space \cite{biber-registervariationonline,sharoff2019}. Not all documents have clear characteristics of one single register, or even any register at all. This has shown also in relatively low inter-annotator agreement for web register annotation \cite{crowston2010problems}.

\begin{table*}[t!]
\centering
\begin{tabular}{lrrrr}
\textbf{General   register category} & \textbf{English} & \textbf{Finnish} & \textbf{French} & \textbf{Swedish} \\ \hline
NA Narrative                         & 36.46 \%         & 34.95 \%         & 22.33 \%        & 28.32 \%         \\
IN Informational description         & 19.24 \%         & 17.03 \%         & 20.74 \%        & 27.68 \%         \\
OP Opinion                             & 16.23 \%         & 15.23 \%         & 6.33 \%         & 6.60 \%          \\
ID Interactive discussion           & 6.77 \%          & 6.29 \%          & 8.03 \%         & 3.57 \%          \\
HI How-to/Instructions               & 3.08 \%          & 6.47 \%          & 3.08 \%         & 2.80 \%          \\
IP Informational persuasion       & 2.75 \%          & 20.04 \%         & 24.15 \%        & 16.82 \%         \\
LY Lyrical                           & 1.32 \%          & 0.00 \%          & 0.33 \%         & 0.14 \%          \\
SP Spoken                            & 1.21 \%          & 0.00 \%          & 0.83 \%         & 0.14 \%          \\
Empty                                & 1.20 \%          & 0.00 \%          & 0.00 \%         & 0.00 \%          \\
Hybrids                              & 11.74 \%         & 0.00 \%          & 14.19 \%        & 13.93 \%         \\ \hline
\textbf{Total}                       & \textbf{48452}   & \textbf{2226}    & \textbf{1818}   & \textbf{2182}   
\end{tabular}
\caption{Proportional register distribution and total number of documents in CORE, FinCORE, FreCORE and SweCORE. Hybrids include all documents annotated with several register labels, and Empty refers to documents not assigned any label.}
\label{tab:regdistribution}
\end{table*}

Very recently, however, the advances brought to NLP by neural networks have shown that registers can be identified also in a corpus featuring the full range of online language variation \cite{laippalalrev}. \newcite{laippala-etal-2019-toward} extended the possibilities of web register identification beyond English by presenting an online register corpus on Finnish (FinCORE) and demonstrating that web registers can be modeled also in a cross-lingual setting.

In this paper, we substantially extend on this early work on cross-lingual web register identification through the following contributions: 1) we introduce manually annotated web register datasets for two new languages, French and Swedish, 2) we demonstrate competitive performance for cross-lingual transfer of a register classification model from English to other languages in a zero-shot setting, and 3) we analyze zero-shot vs.\ monolingual training for register classification and remaining challenges in both. In particular, using Transformer-based pre-trained language models, we show that a zero-shot cross-lingual approach outperforms monolingual results achieved by a previously proposed state-of-the-art method for all the three language pairs (En-Fr, En-Sv, and En-Fi), and that strong monolingual performance can be achieved with limited training data.

\section{Data}
\label{sec:data}

\begin{table*}[h]
\centering
\begin{tabular}{l|rr|rr|rr|rr|}
\multicolumn{1}{c|}{Register} & \multicolumn{2}{c|}{English} & \multicolumn{2}{c|}{Finnish} & \multicolumn{2}{c|}{French} & \multicolumn{2}{c|}{Swedish} \\
% & mean & std. & mean & std. & mean & std. & mean & std. \\ \hline
  & \multicolumn{1}{c}{mean} & \multicolumn{1}{c|}{std.} & \multicolumn{1}{c}{mean} & \multicolumn{1}{c|}{std.} & \multicolumn{1}{c}{mean} & \multicolumn{1}{c|}{std.} & \multicolumn{1}{c}{mean} & \multicolumn{1}{c|}{std.} \\ \hline
NA & 1081 & 2490 & 649 & 2170 & 623 & 2284 & 602 & 2461 \\
IP & 1066 & 3370 & 301 & 391 & 325 & 493 & 426 & 2225 \\
IN & 1353 & 3373 & 989 & 4755 & 1446 & 9688 & 323 & 626 \\
OP & 1595 & 4021 & 739 & 1188 & 857 & 1835 & 1055 & 1825 \\
HI & 1007 & 1402 & 277 & 285 & 623 & 1130 & 437 & 508 \\
ID & 1079 & 4042 & 2017 & 8907 & 970 & 1579 & 577 & 885 \\
LY & 468 & 1114 & \multicolumn{1}{c}{-} & \multicolumn{1}{c|}{-} & 387 & 314 & 263 & 225 \\
SP & 2047 & 3335 & \multicolumn{1}{c}{-} & \multicolumn{1}{c|}{-} & 999 & 939 & 525 & 178 \\
Empty & 13345 & 3215 & \multicolumn{1}{c}{-} & \multicolumn{1}{c|}{-} & \multicolumn{1}{c}{-} & \multicolumn{1}{c|}{-} & \multicolumn{1}{c}{-} & \multicolumn{1}{c|}{-}\\
Hybrids & 1290 & 3141 & \multicolumn{1}{c}{-} & \multicolumn{1}{c|}{-} & 1170 & 3296 & 859 & 1207 \\\hline
All & 1083 & 2747 & 713 & 3295 & 703 & 3900 & 482 & 1446
\end{tabular}
\caption{Average length (number of words) and standard deviation of Finnish, French, Swedish and English documents.}
\label{tab:document-length}
\end{table*}

We use four register-annotated corpora representing the unrestricted open web: the English CORE and Finnish FinCORE, which have been introduced in previous work \cite{egbert:core,laippala-etal-2019-toward}, and two new corpora, FreCORE for French and SweCORE for Swedish. These novel datasets are released under open licences together with this paper.\footnote{Available at \url{https://github.com/TurkuNLP/Multilingual-register-corpora}} With these new resources, the possibilities for web register identification expand substantially.

FreCORE and SweCORE are random samples of the 2017 CoNLL datasets \cite{ginter2017conll} originally drawn from Common Crawl. Both datasets were deduplicated using Onion \cite{pomikalek2011removing} with 0.7 threshold and n-gram length of 5. 
All material not belonging to the body of text, such as boilerplate, was removed. Titles, however, were preserved. The cleaning and pre-processing steps follow the procedure suggested in \newcite{laippala2020web}. The register annotation of the datasets was conducted individually by two trained annotators with a linguistics background. Uncertain cases were discussed and resolved together with an annotation supervisor. The inter-annotator agreement, counted prior to the discussions, was 78\% F1-score for FreCORE and 84\% for SweCORE. This can be considered as a lower bound.

All datasets are similarly annotated across languages, and they all apply the same hierarchical register class taxonomy originally introduced for CORE. It includes eight main registers (e.g., Narrative) and approximately 30 sub-registers (e.g., News report within Narrative). The main and sub-register categories are illustrated in the appendix. %Appendix \ref{sec:appendix}
 When a document shares characteristics of several registers, it can be assigned several labels both at the main and sub-register level. These documents are called \textit{hybrids}.
As our focus in this paper is on general register categories, we initially pre-process all four corpora to remove the more specific sub-register labels.

The general register categories and their distributions as well as the average document length and standard deviation for all classes are presented in Table \ref{tab:regdistribution} and Table \ref{tab:document-length}, respectively. 
The register class Empty consists of texts whose register the annotators could not agree on. Due to the very small number of each type of hybrid label combination in the data, in Tables \ref{tab:regdistribution} and \ref{tab:document-length}, the class Hybrids includes all documents that have more than one label. Table \ref{tab:regdistribution} reveals that the register distributions in the four languages are broadly similar, featuring Narrative, Informational description, and hybrids among the four most frequent categories. The top four also include Informational persuasion in FinCORE, FreCORE, and SweCORE, while in CORE this label is relatively infrequent. Additionally, Opinion is notably more frequent in CORE and FinCORE than in FreCORE and SweCORE. These differences may reflect differences in data compilation. Table \ref{tab:document-length} shows that, on average, English documents are longer than documents in other languages, whereas Swedish documents tend to be shortest. Overall the number of words in a document in most of the classes show large variation, with the longest documents containing tens of thousands of words.

\section{Experimental setup}
\label{sec:models}

The architectures and models we are using are presented below.\footnote{The code is available at: \url{https://github.com/TurkuNLP/Multilingual-register-corpora}} We perform multi-label document classification, where each document can have zero, one, or several register labels.
The experiments are divided into 1) a monolingual setup with training and evaluation on Finnish, French, Swedish, and English (as reference), and 2) a zero-shot cross-lingual setup with training on English and evaluation on the other languages.

\textbf{BERT}, Bidirectional Encoder Representations from Transformers \cite{devlin2019bert} is a state-of-the-art deep bidirectional language model pre-trained on large unlabelled corpora. BERT's architecture is a multi-layer Transformer encoder that is based on the original Transformer architecture introduced by \newcite{vaswani2017attention}.
We use cased BERT models (TensorFlow versions) %large size when available)
through the Huggingface Transformers library \cite{wolf2020transformers} with the following language-specific models: the original English BERT, Finnish FinBERT \cite{virtanen2019multilingual}, French FlauBERT \cite{le2020flaubert} and Swedish KB-BERT \cite{malmsten2020playing}. Additionally, we use Multilingual BERT (mBERT) \cite{devlin2019bert}, which was pre-trained on monolingual Wikipedia corpora from 104 languages with a shared multilingual vocabulary.

\textbf{XLM-RoBERTa} (XLM-R, \newcite{conneau2020unsupervised}) is a multilingual language model that follows the Cross-lingual Language Modeling (XLM) approach \cite{conneau2019cross} and is based on the RoBERTa model \cite{liu2019roberta}, which shares the architecture of BERT. The authors argue that XLM and mBERT are undertuned and that the improved and prolonged training procedure of RoBERTa in combination with more data -- on average two orders of magnitude more for low-resource languages -- is key to improving cross-lingual performance. XLM-R is trained on 2.5TB of filtered Common Crawl \cite{wenzek2019ccnet} data comprising of monolingual texts in 100 languages. 
It is claimed to be the first multilingual model to outperform monolingual models, as well as Multilingual BERT in a number of experiments \cite{conneau2020unsupervised,libovicky2020language,tanase2020upb}. 

\begin{table*}[ht!]
%\centering
%\small
\setlength{\tabcolsep}{0.4em}
\begin{tabular}[t]{l|lllll}
&\multicolumn{5}{c}{\textbf{Monolingual}}\\
Model & Train- &  \multicolumn{2}{c}{Dev} & \multicolumn{2}{c}{Test} \\  %& \multicolumn{2}{c}{ROC-AUC (SD)} \\
 & Test & F1 (\%) & Std. & F1 (\%) & Std. \\\hline %& Dev & Test \\ \hline

CNN & Fi & 59.04 & (0.67) & 58.04 & (1.02) \\ %& 0.872 (0.013) & 0.872 (0.005) \\
%Fi & FinBERT & \textbf{0.78}  &  &  &  \\
 %&	0.9365 (0.0037) & \\
%Fi & WikiBERT-Fi & 0.679 (0.019) & 0.633 (0.010) & 0.948 (0.005) & 0.934 (0.002) \\
%Fi & biBERT & 0.752 (0.008) & 0.699 (0.005) & 0.960 (0.003) & 0.941 (0.006) \\
mBERT & Fi & 65.91 &(0.85) & 64.83 &(1.16) \\ %&	0.8975 (0.0024) & \\
%Fi & XLM-R (base) & 0.7274 (0.0110) & 0.7115 (0.0153) & 0.9317 (0.0014) &  \\
XLM-R large& Fi &  76.25 &(0.45) & \textbf{73.18} &(1.35)\\
FinBERT & Fi & \textbf{76.28} &(1.23) & 72.98 & (0.74) \\\hline % & 0.9413 (0.0057)  &  \\ \hline
CNN & Fr & 59.78 & (1.10) & 58.14 &(1.10) \\ %& 0.892 (0.004) & 0.890 (0.002)  \\
%& 0.8825 (0.0087) &  \\
%Fr & WikiBERT-Fr & 0.693 (0.015) & 0.681 (0.013)  & 0.934 (0.008) & 0.921 (0.008) \\
mBERT & Fr & 70.74 &(1.67) & 68.66 &(0.63) \\ %& 0.9167 (0.0047) & \\
%Fr & XLM-R (base) & 0.7419 (0.0022) & 0.7307 (0.0108) & 0.9235 (0.0073) & \\
XLM-R large & Fr & \textbf{77.38} &(0.51) & \textbf{76.92} &(0.24) \\
FlauBERT large&  Fr & 73.93 &(0.93) & 72.56 &(1.40) \\\hline %& 0.9432 (0.0077) &  \\ \hline

CNN & Sv & 69.43 &(0.56) & 67.89 &(1.01) \\ %& 0.933 (0.000) & 0.926 (0.003)  \\

%Sv & WikiBERT-Sv & 0.762 (0.010) & 0.744 (0.012) & 0.954 (0.006) & 0.949 (0.000) \\
mBERT & Sv & 76.91 &(0.45) & 76.43 &(0.46) \\ %& 0.8307 (0.0038) & \\
%Sv & XLM-R (base) & 0.7915 (0.0110)	& 0.8058 (0.0143) & 0.8401 (0.0246) & \\
%0.8017 (0.0084) &  & 0.8588 (0.0405) &  \\
XLM-R large & Sv & \textbf{82.61} &(0.37) & \textbf{83.04} &(0.62)\\
KB-BERT & Sv & 80.15 & (0.50) & 80.75 & (0.09) \\ 
\hline 
%\\\hline %& 0.8879 (0.0132)  &  \\ \hline

CNN & En & 64.56 & (0.78)  & 64.03 & (0.30) \\%& 0.925 (0.002) & 0.927 (0.002)  \\
%En & BERT (base) & 0.721 (0.002) & 0.721 (0.002) & 0.955 (0.000) & 0.958 (0.000) \\
% & 0.958 (0.000) & 0.959 (0.001) \\
%En & WikiBERT-En & 0.716 (0.001) & 0.715 (0.000) & 0.956 (0.000) & 0.956 (0.001) \\
%En & biBERT & 0.728 (0.000) & 0.728 (0.003) & 0.958 (0.000) & 0.960 (0.000) \\
mBERT & En & 72.80 &(0.21) & 73.06 &(0.09) \\%& 0.954 (0.000) & 0.957 (0.000) \\
%En & XLM-R (base) & 0.7442 (0.0019)	& 0.7475 (0.0012) & 0.9336 (0.0002) &  \\
XLM-R large & En & \textbf{75.80} &(0.12)  & \textbf{75.68} &(0.05)\\
BERT large & En & 74.01 & (0.42) & 74.07 &(0.28)\\ %& 0.9421 (0.0010)  &  \\ \hline

\end{tabular}
\begin{tabular}[t]{|lllll}
\multicolumn{5}{c}{\textbf{Cross-lingual}}\\
Train-  & \multicolumn{2}{c}{Dev} & \multicolumn{2}{c}{Test} \\
Test &  F1 (\%) & Std. & F1 (\%) & Std. \\\hline %& Dev & Test \\ \hline
En-Fi & %CNN &
40.53 &(1.11) & 41.56 &(0.20) \\ % & 0.825  (0.001) & 0.824 (0.002) \\
%En-Fi & biBERT & 0.574 (0.007) & 0.526 (0.023) & 0.906 (0.003) & 0.880 (0.015) \\
En-Fi &% mBERT & 
51.02 &  (2.92) & 50.21 & (0.74) \\ %& 0.881 (0.008) & 0.873 (0.007) \\
%En-Fi & XLM-R (base) & 0.5788 (0.0158) & 0.5803 (0.0015) & 0.6721 (0.0045) & \\
En-Fi & %XLM-R (L) & 
\textbf{61.60} &(2.01) & \textbf{61.35} &(1.26) \\ %& 0.6652 (0.0050) &  \\ \hline
%\multicolumn{5}{c}{\cellcolor{gray!10}}\\ 
\\
\hline
%\rowcolor{Grey}
%&&&&\\\hline 
En-Fr & %CNN &  
46.44 &(0.51) & 46.78 &(1.80) \\ % & 0.842 (0.002) & 0.841 (0.003) \\
En-Fr & %mBERT & 
56.73 & (1.54) & 55.04 &(0.66) \\ %& 0.901 (0.002) & 0.903 (0.002) \\
%En-Fr & XLM-R (base) & 0.6430 (0.0116) &&	0.9323 (0.0024)  &  \\
En-Fr & %XLM-R (L) & 
\textbf{65.66} &(0.52)	& \textbf{64.27} &(1.58) \\%& 0.9246 (0.0060) &  \\ 
%\multicolumn{5}{c}{\cellcolor{gray!10}}\\ 
%\rowcolor{Grey}
%&&&&\\
\\
\hline
En-Sv & %CNN & 
43.74 &(0.82) & 43.78 &(1.00) \\ %& 0.835 (0.010) & 0.832 (0.005)  \\
En-Sv & %mBERT & 
62.37 &(0.82) & 62.53 & (0.78) \\ %& 0.905 (0.010) & 0.891 (0.010) \\
%En-Sv & XLM-R (base) & 0.7035 (0.0214) & & 0.9112 (0.0275) & \\
En-Sv & %XLM-R (L) & 
\textbf{70.49} &(0.58) & \textbf{69.22} &(1.66) \\ %& 0.8947 (0.0290) & 
%\multicolumn{5}{c}{\cellcolor{gray!10}}\\ 
%\rowcolor{Grey}
%&&&&\\
\\
\hline

\end{tabular}
\caption{Monolingual and zero-shot cross-lingual classification results (N=3). Best results for each experiment shown in bold.}
\label{tab:all-results}
\end{table*}

We also apply a \textbf{CNN} (Convolutional Neural Network) based architecture following \newcite{kim2014convolutional}, as our baseline model. We modify the cross-lingual CNN used by \newcite{laippala-etal-2019-toward} to a multi-label setting. We use the multilingual word vectors introduced by \newcite{Conneau2017word}.
The CNN employs a convolution layer with ReLU activation, a max-pooling layer and sigmoid activation.

The French and Swedish data were divided into training, development and test sets using stratified sampling with a 50/20/30 split. For BERT-based models we used large model size when available to maximize model performance. We used the maximum sequence length of 512 tokens (with truncation at the end) and batch size of 7, and performed a grid search on learning rate (8e\textsuperscript{-6}--6e\textsuperscript{-5}) and number of training epochs (3--7). For the CNN, we performed a grid search on the kernel size (1--2), learning rate (1e\textsuperscript{-4}--1e\textsuperscript{-2}), and prediction threshold (0.4, 0.5, 0.6).

\section{Results}
\label{sec:results}

In Table \ref{tab:all-results}, we present the primary results on English, Finnish, French and Swedish monolingual classification with the models described in Section \ref{sec:models}, as well as cross-lingual results with English as the source language and Finnish, French and Swedish as target languages. We report the mean and standard deviation of F1 over three repetitions.

In monolingual settings, XLM-R large performs competitively compared to monolingual models and clearly outperforms both mBERT and the CNN baseline. The lead of XLM-R over monolingual models is substantial in all cases except for the FinBERT model, where the two perform within one standard deviation of each other. 
Our results support the claimed competitiveness of XLM-R large with monolingual models, mentioned in Section \ref{sec:models}. 

\begin{figure}[t]
\includegraphics[width=\linewidth,clip,trim={0.25cm 0cm 0.2cm 0cm}]{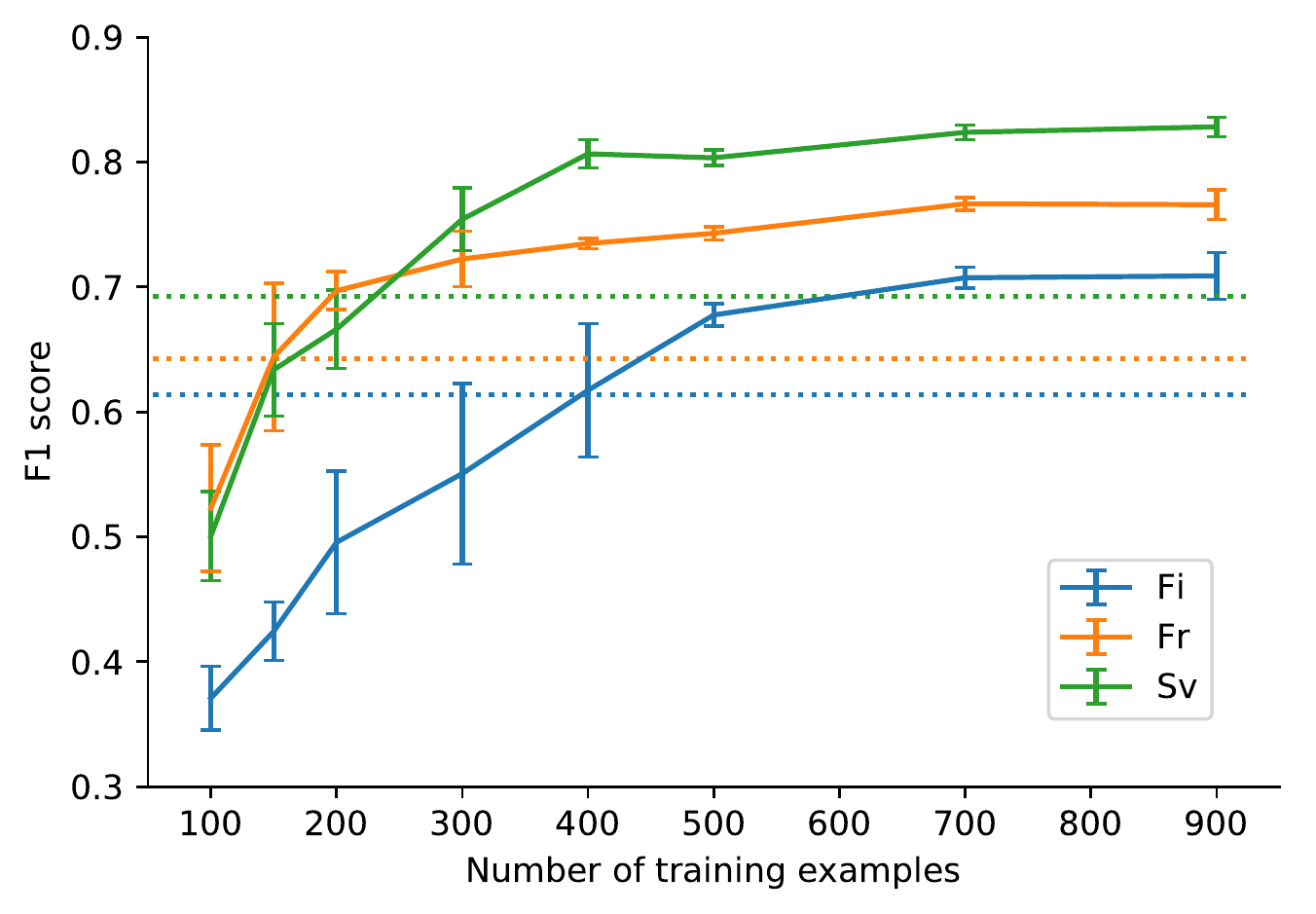}
%\setlength{\belowcaptionskip}{-15pt}
%\captionsetup[figure]{font=small,skip=0pt}
\caption{Monolingual performance when training with varying number of examples (solid lines) in relation to zero-shot cross-lingual performance when training on full English set (dotted lines). Error bars represent standard deviations (N=6).}
\label{fig:monolingual_light}
\end{figure}

\begin{figure*}[ht!]
\centering
% trim order is left bottom right top
%\includegraphics[trim={2cm 4cm 8cm 4cm},clip,scale=0.25]{figs/en-fr} 
%\includegraphics[trim={2cm 4cm 8cm 4cm},clip,scale=0.25]{figs/en-sv-set.pdf
\includegraphics[trim={0cm 0cm 0.5cm 0.3cm},clip,scale=0.175]{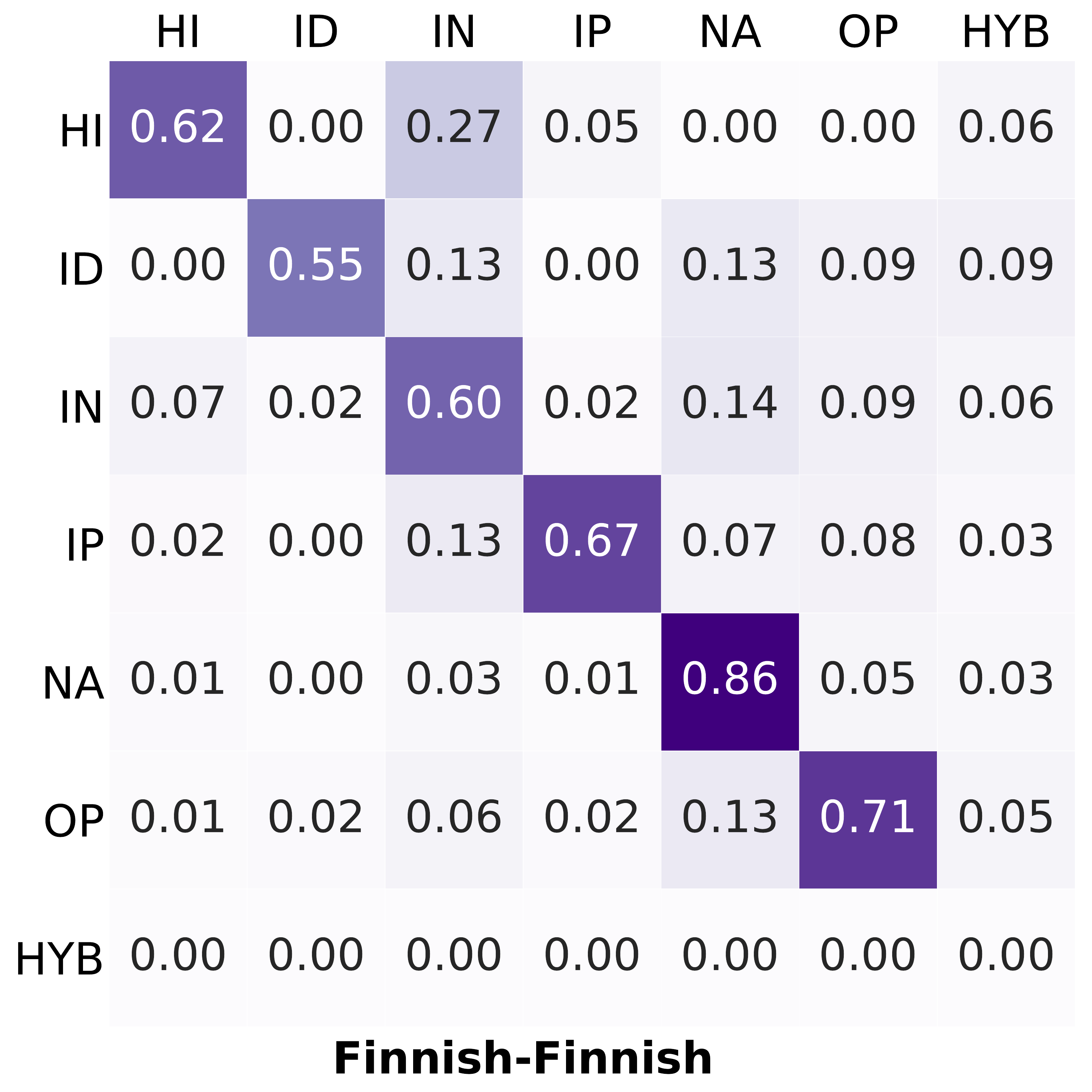}
\includegraphics[trim={0cm 0cm 0.5cm 0.3cm},clip,scale=0.175]{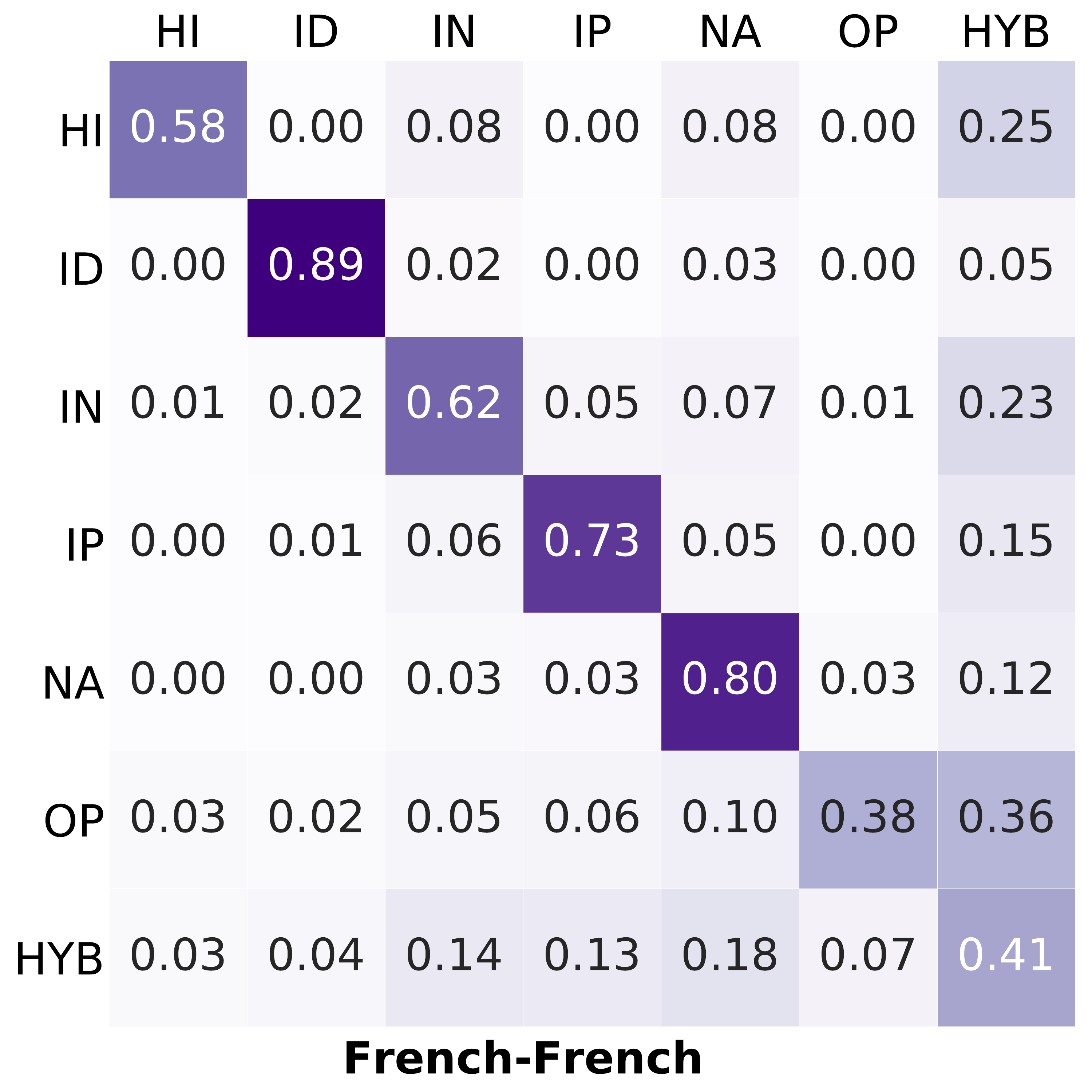}
\includegraphics[trim={0cm 0cm 0.5cm 0.3cm},clip,scale=0.175]{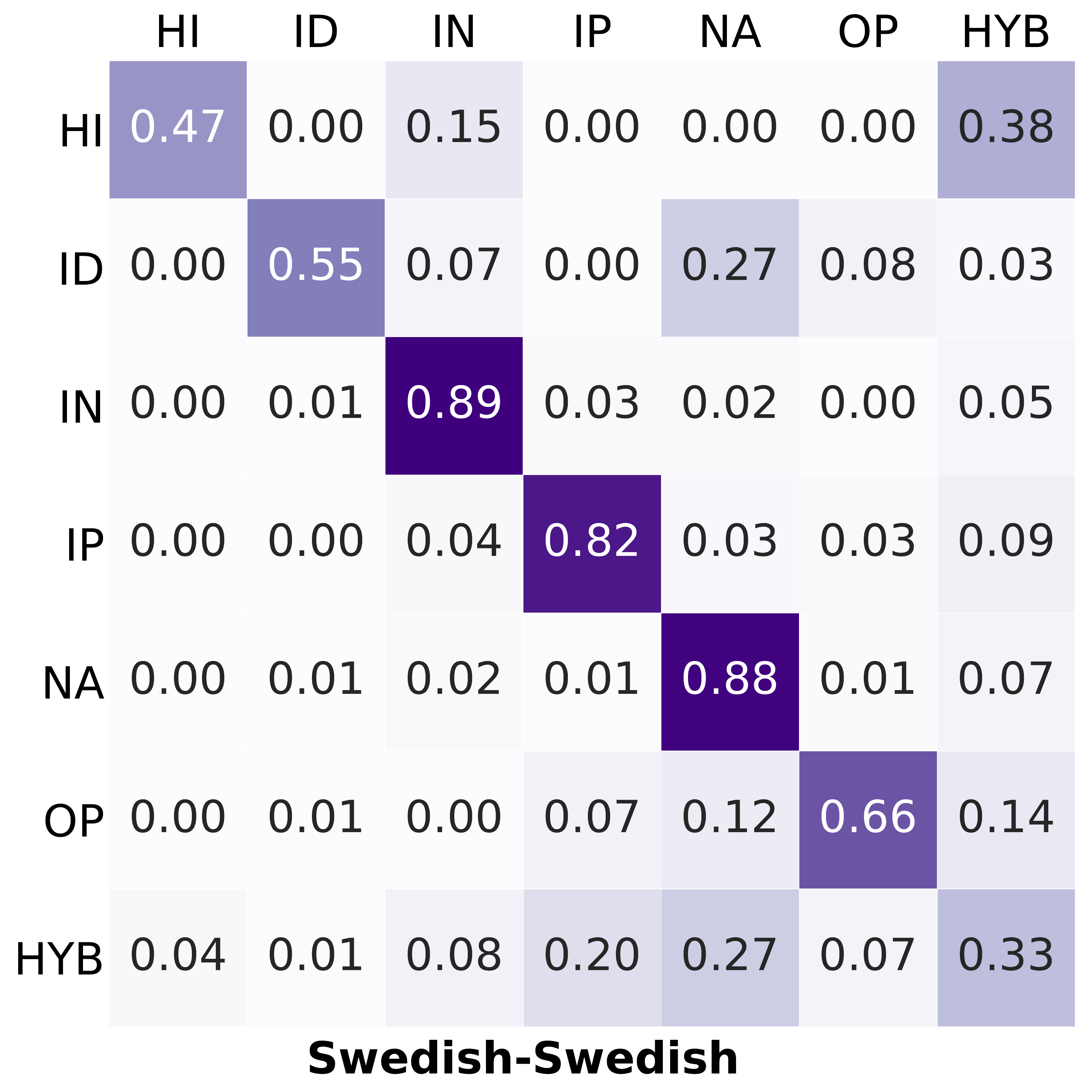}
% zeroshot
\includegraphics[trim={0cm 0cm 0.5cm 0.0cm},clip,scale=0.175]{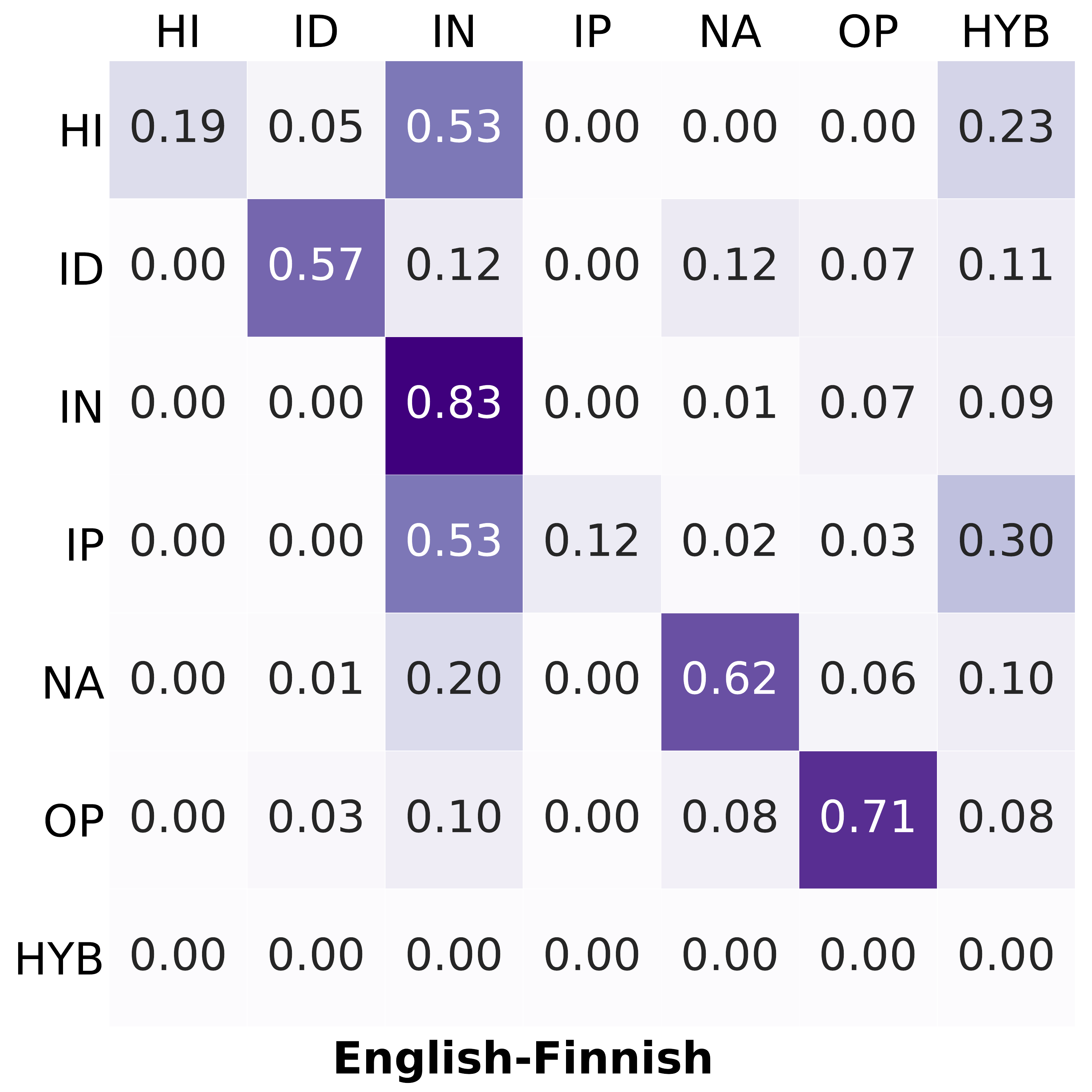}
\includegraphics[trim={0cm 0cm 0.5cm 0.0cm},clip,scale=0.175]{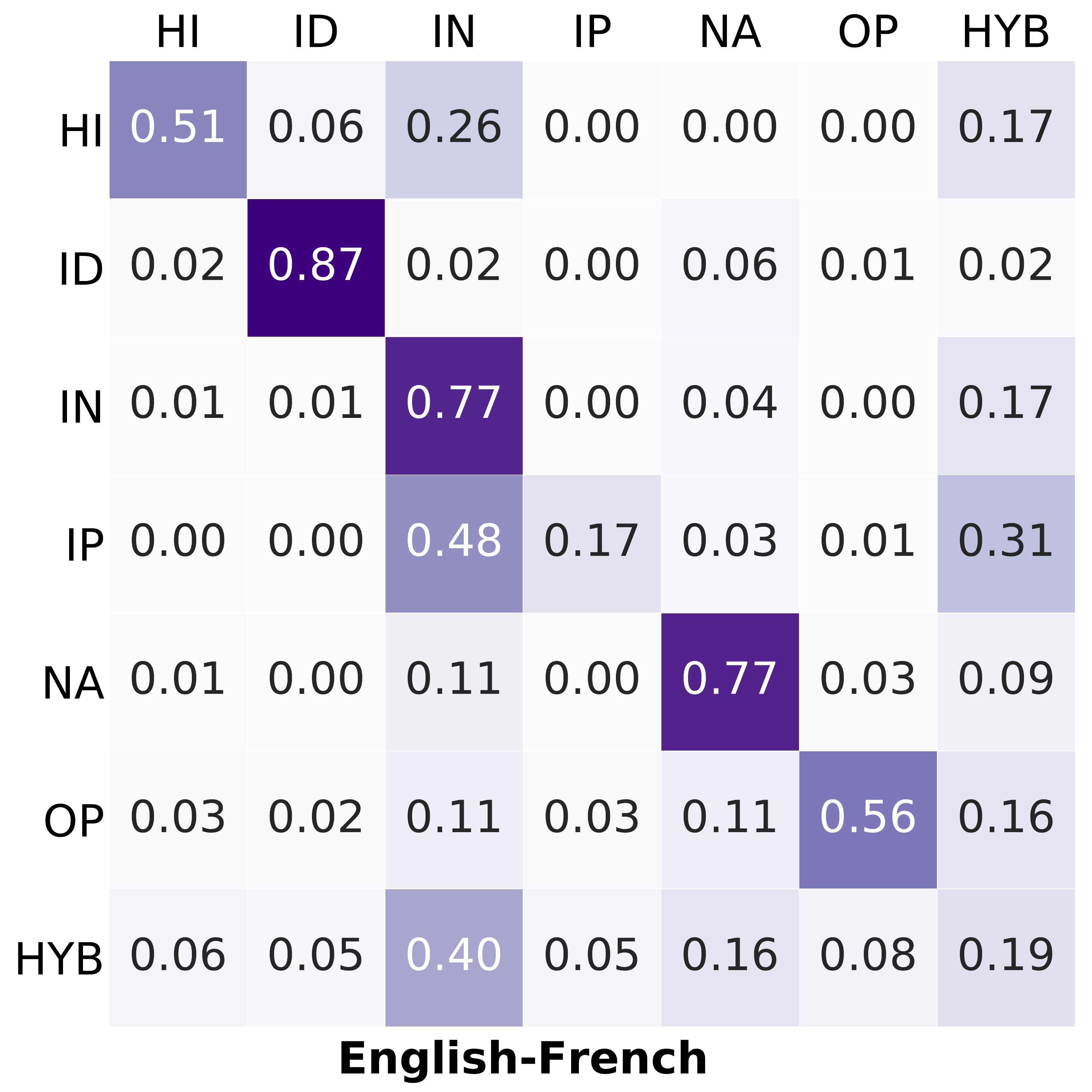}
\includegraphics[trim={0cm 0cm 0.5cm 0.0cm},clip,scale=0.175]{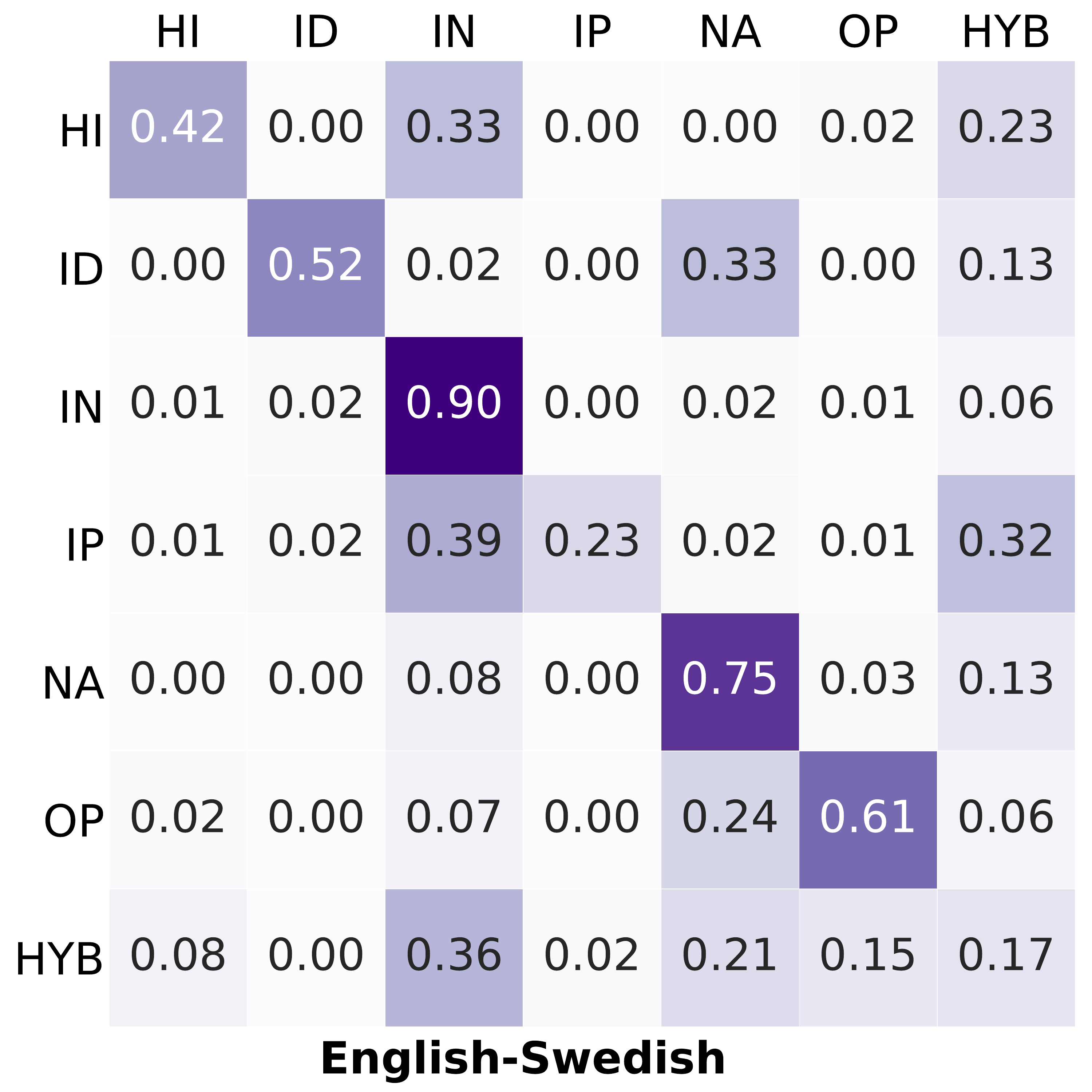}

\caption{Confusion matrices presenting class label observations (rows) vs. class label predictions (columns) in monolingual (upper row) and cross-lingual (lower row) settings. The numbers and coloring represent the proportions of predictions per row. HYB is a combination of all hybrid cases with multiple labels.}
\label{fig:conf_matrix}
\end{figure*}

English, Finnish and French BERT models achieve similar monolingual test results (73--74\% F1-score), while the Swedish KB-BERT achieves the highest F1-score (81\%). The Finnish classification task is seemingly easier due to smaller number of classes, nevertheless, other factors may cause the difficulty of the task to differ between languages. For instance, the measured human inter-annotator agreements at 78\% (Fr) and 84\% (Sv) F1-score (see Section~\ref{sec:data}) represent a theoretical upper bound for the classification task and reflect the tendency of Swedish being easier to classify; the level of agreement has not been reported for Finnish.
Although not strictly comparable, our results clearly outperform the previous state-of-the-art results achieved with the CNN \cite{laippala-etal-2019-toward} in terms of F1, which in turn outperforms \newcite{biber2016-detecting}, who used the same corpus but in multi-class setting.

Furthermore, Table \ref{tab:all-results} shows very strong zero-shot cross-lingual results with XLM-R large, with F1-scores in the 61--69\% range. This represents a remarkably consistent relative decrease of 16.2--16.6\% (11.8--13.8\% absolute) from the monolingual scores of XLM-R. 
Its lead over mBERT increases from 6.6--8.4\% absolute F1 to 7.8--11.4\% in the cross-lingual settings, whereas its lead over the CNN goes from 15.1--18.8\% to 17.5--25.4\%. Most interestingly, the zero-shot XLM-R even beats the monolingually trained CNN baselines by a significant margin for Finnish and French, while its lead remains within a standard deviation for Swedish.

In Figure \ref{fig:monolingual_light}, we illustrate the effect of training monolingual XLM-R large models with varying train set sizes and compare the performance against the reported zero-shot performance. The optimal monolingual hyperparameter settings for each language are used, while training the model instances on 100--900 examples each. We see that zero-shot cross-lingual performance is surpassed already with about 150 training instances for French, 225 for Swedish and 400 for Finnish, while performance seems to converge around 500.

Previous studies have shown repeatedly that registers vary considerably in terms of how well they are linguistically defined and thus how well they can be automatically identified \cite{biber-registervariationonline,biber-core-mda,laippalalrev}. For instance, while texts in the IN (Informational description) and NA (Narrative) classes, such as Encyclopedia articles and Sports reports, have very distinctive characteristics and can be identified with a very high reliability, others, such as Information blogs in the IN class or Advice in the OP (Opinion) class receive much lower scores.

Figure \ref{fig:conf_matrix} presents confusion matrices on the predictions in monolingual and cross-lingual settings, using the best-performing model.\footnote{See appendix for class-specific F1 results.} For the sake of simplicity, the multi-label predictions have been collapsed into multi-class by including all hybrids under one label HYB in Figure \ref{fig:conf_matrix}. In the monolingual settings, we can see that particularly hybrids present a challenge. This is expected, as they feature characteristics of several registers. Additionally, while IP (Informational persuasion) and NA are predicted with high performance in all three languages, the other classes display more variation. For instance, ID (interactive discussion) reaches an F1-score of 90\% (see appendix) in French monolingual setting, whereas in Swedish and Finnish it is frequently misclassified, most likely because of the small number of examples in the training data.

The hybrids are also frequently misclassified in cross-lingual settings. Interestingly, register classes also feature clear differences in the extent to which the cross-lingual transfer affects the identification performance. The register class IN tends to be predicted strongly in all zero-shot language pairs. This is probably due to the IN class including documents with strong cross-lingual signals. For instance, IN includes Encyclopedia articles (see appendix), such as Wikipedia texts, that tend to be very similar across languages.

While most of the non-hybrid classes experience a small drop in performance, the identification rate for IP and HI (How-to/Instructions) drops dramatically in cross-lingual settings in all language pairs. 
The decrease of IP can be linked to its smaller proportion in the English data (see Section \ref{sec:data}), but the drops experienced by IP and HI can also reflect the variation displayed by registers across languages. \newcite{biber2014using} showed that registers, such as spoken texts, display functional similarities across languages, which obviously is needed for high-quality transfer in register identification. However, analyzing the English CORE registers, \newcite{laippalalrev} noted that some registers, such as many blogs, depend highly on lexical characteristics reflecting the discussion topics. These topics, however, may vary extensively between languages. This, again, may complicate the transfer learning for these classes.

\section{Discussion and conclusions}

Despite the many opportunities that reliable recognition of text register would introduce for the analysis and use of web documents and many efforts to address this task over the years, only limited progress has been made toward unrestricted web document register classification. Previous work has also focused almost exclusively on English.

In this study, we have introduced manual register annotation compatible with that of the large English CORE corpus for two languages previously lacking such a resource, namely French and Swedish. We also demonstrated that state-of-the-art multilingual neural language models can support zero-shot transfer of register annotations from English to a Germanic, Romance and Finnic language at levels of performance broadly comparable or better to previously published monolingual results on CORE.

Moreover, we demonstrated that small amounts of monolingual training data are needed to reach or surpass this level of performance, which attests that reliable register identification in a new language is readily attainable using current pre-trained language models. We further compared and analysed the results for monolingual and cross-lingual register classifiers, finding that certain registers as well as hybrid texts combining several register characteristics continue to pose challenges in particular for cross-lingual transfer. In future work, we will build on these results to extend multi- and cross-lingual modeling in order to create massive multilingual register-annotated web corpora.

 \section*{Acknowledgments}

We thank for the financial support of the Emil Aaltonen Foundation and Academy of Finland. We also wish to acknowledge CSC – IT Center for Science, Finland, for computational resources.

\bibliography{anthology,eacl2021_2}
\bibliographystyle{acl_natbib}

\newpage

\appendix

\begin{table*}[]
\centering
\begin{tabular}{l|rr|rr|rr}
            & \multicolumn{2}{c|}{\textbf{En-Fi}}                                  & \multicolumn{2}{c|}{\textbf{En-Fr}}                                  & \multicolumn{2}{c}{\textbf{En-Sv}}                                  \\
            & \multicolumn{1}{c}{\textbf{F1}} & \multicolumn{1}{c|}{\textbf{Std.}} & \multicolumn{1}{c}{\textbf{F1}} & \multicolumn{1}{c|}{\textbf{Std.}} & \multicolumn{1}{c}{\textbf{F1}} & \multicolumn{1}{c}{\textbf{Std.}} \\ \hline
\textbf{HI} & 48.43 \%                        & 1.98 \%                            & 55.12 \%                        & 5.65 \%                            & 62.91 \%                        & 0.60 \%                           \\
\textbf{ID} & 69.79 \%                        & 6.06 \%                            & 87.48 \%                        & 2.05 \%                            & 52.05 \%                        & 3.07 \%                           \\
\textbf{IN} & 44.43 \%                        & 0.32 \%                            & 58.68 \%                        & 0.17 \%                            & 68.81 \%                        & 0.20 \%                           \\
\textbf{IP} & 52.79 \%                        & 5.72 \%                            & 53,57 \%                        & 2.53 \%                            & 51.45 \%                        & 1.88 \%                           \\
\textbf{LY} & 0.00 \%                         & 0.00 \%                            & 66.67 \%                        & 0.00 \%                            & 95.24 \%                        & 6.73 \%                           \\
\textbf{NA} & 77.85 \%                        & 0.80 \%                            & 75.18 \%                        & 0.32 \%                            & 78.36 \%                        & 0.70 \%                           \\
\textbf{OP} & 70.32 \%                        & 1.59 \%                            & 59.26 \%                        & 1.51 \%                            & 60.57 \%                        & 0.32 \%                           \\
\textbf{SP} & 0.00 \%                         & 0.00 \%                            & 79.08 \%                        & 7.53 \%                            & 0.00 \%                         & 0.00 \%                          
\end{tabular}
\caption{Class-wise F1-scores and standard deviations on cross-lingual experiments}
\label{tbl:class-wise-zero-shot-f1}
\end{table*}

\begin{table*}[]
\centering
\begin{tabular}{l|rr|rr|rr}
            & \multicolumn{2}{c|}{\textbf{Fi-Fi}}                                  & \multicolumn{2}{c|}{\textbf{Fr-Fr}}                                  & \multicolumn{2}{c}{\textbf{Sv-Sv}}                                  \\
            & \multicolumn{1}{c}{\textbf{F1}} & \multicolumn{1}{c|}{\textbf{Std.}} & \multicolumn{1}{c}{\textbf{F1}} & \multicolumn{1}{c|}{\textbf{Std.}} & \multicolumn{1}{c}{\textbf{F1}} & \multicolumn{1}{c}{\textbf{Std.}} \\ \hline
\textbf{HI} & 64.02 \%                        & 1.94 \%                            & 58.81 \%                        & 0.59 \%                            & 70.70 \%                        & 4.56 \%                           \\
\textbf{ID} & 66.18 \%                        & 3.54 \%                            & 90.37 \%                        & 1.58 \%                            & 60.48 \%                        & 3.21 \%                           \\
\textbf{IN} & 58.68 \%                        & 1.59 \%                            & 74.00 \%                        & 0.40 \%                            & 87.79 \%                        & 0.29 \%                           \\
\textbf{IP} & 75.74 \%                        & 2.34 \%                            & 80.02 \%                        & 1.04 \%                            & 81.75 \%                        & 1.10 \%                           \\
\textbf{LY} &                                 -- &                                    -- & 66.67 \%                        & 0.00 \%                            & 0.00 \%                         & 0.00 \%                           \\
\textbf{NA} & 82.38 \%                        & 0.98 \%                            & 77.02 \%                        & 1.16 \%                            & 86.66 \%                        & 0.69 \%                           \\
\textbf{OP} & 67.10 \%                        & 2.05 \%                            & 66.23 \%                        & 3.08 \%                            & 75.37 \%                        & 1.66 \%                           \\
\textbf{SP} &                               --  &                                  --  & 65.28 \%                        & 1.96 \%                            & 0.00 \%                         & 0.00 \%                          
\end{tabular}
\caption{Class-wise F1-scores and standard deviations on monolingual experiments}
\label{tbl:class-wise-monolingual-f1}
\end{table*}

\section{Appendix}
\label{sec:appendix}

Tables~\ref{tbl:class-wise-zero-shot-f1} and~\ref{tbl:class-wise-monolingual-f1} present the detailed results for the zero-shot cross-lingual and monolingual register classification experiments, respectively. Table ~\ref{tbl:registers-all} presents the register taxonomy with the main registers and their sub-registers.

\begin{table*}[]
%\small
\setlength\tabcolsep{3pt} % default value: 6pt
\centering
\begin{tabular}{p{5pt}l}
\multicolumn{2}{l}{\textbf{Narrative}} \\
& \multirow{3}{0.45\textwidth}{News report / news blog, sports report, personal blog, historical article, fiction, travel blog, community blog, online article} \\ \\ \\
\multicolumn{2}{l}{\textbf{Informational description}} \\
& \multirow{4}{0.45\textwidth}{Description of a thing, encyclopedia article, research article, description of a person, information blog, FAQ, course material, legal terms / condition, report, job description} \\ \\ \\ \\ 
\multicolumn{2}{l}{\textbf{Opinion}} \\
& Review, opinion blog, religious blogs/sermon, advice \\
\multicolumn{2}{l}{\textbf{Interactive discussion}} \\
& Discussion forum, question-answer forum \\
\multicolumn{2}{l}{\textbf{How-to/Instructions}} \\
& How-to/instruction, recipe\\
\multicolumn{2}{l}{\textbf{Informational Persuasion}} \\
& \multirow{2}{0.45\textwidth}{Description with intent to sell, news+opinion blog / editorial} \\ \\
\multicolumn{2}{l}{\textbf{Lyrical}} \\
& Songs, poem\\
\multicolumn{2}{l}{\textbf{Spoken}} \\
& Interview, formal speech, TV transcript\\ 
\end{tabular}
\caption{All register classes. Main registers are shown in bold.}
\label{tbl:registers-all}
\end{table*}

\end{document}